\RequirePackage[svgnames,table]{xcolor}

\documentclass[11pt,letterpaper,logo]{yalearxiv}

\usepackage{graphicx}
\usepackage{float,epstopdf}
\usepackage{bbm}

\usepackage{microtype}

\usepackage{natbib}
\setcitestyle{square}

\usepackage{subcaption}
\usepackage{booktabs}

\usepackage{amsmath}
\usepackage{amssymb}
\usepackage{mathtools}
\usepackage{amsthm}
\usepackage{dsfont}
\usepackage{multicol}
\usepackage{makecell}
\usepackage{multirow} 
\usepackage{amsfonts} 
\usepackage{mathrsfs}
\usepackage[amssymb, thickqspace]{SIunits}
\usepackage{enumitem}
\usepackage{pgfplotstable}
\usepackage{lipsum}		

\usepackage{microtype}
\usepackage{graphicx}
\usepackage{booktabs} 
\usepackage[table]{xcolor}
\usepackage{arydshln}
\usepackage[normalem]{ulem} 

\usepackage{cases}
\usepackage{wrapfig}

\usepackage{url}

\usepackage{thmtools}
\usepackage{thm-restate}
\usepackage{tabu}

\definecolor{huskypurple}{HTML}{4B2E83}

\usepackage{titletoc}

\usepackage{listings}
\lstdefinestyle{promptstyle}{
  basicstyle=\ttfamily\footnotesize,
  breaklines=true,
  breakautoindent=false,
  breakindent=0pt,
  postbreak=\mbox{\textcolor{gray}{$\hookrightarrow$}\space},
  columns=fullflexible,
  keepspaces=true,
  frame=single,
  framesep=5pt,
  xleftmargin=6pt,
  xrightmargin=6pt,
  aboveskip=8pt,
  belowskip=8pt,
  showstringspaces=false,
}   
\makeatletter
\def\munderbar#1{\underline{\sbox\tw@{$#1$}\dp\tw@\z@\box\tw@}}
\makeatother

\AddToHook{cmd/appendix/before}{%
  \setcounter{axiom}{0}%
}

\newcommand{\be}{\begin{equation}}
\newcommand{\ee}{\end{equation}}
\newcommand{\bea}{\begin{equation*}\begin{aligned}}
\newcommand{\eea}{\end{aligned}\end{equation*}}

\newtcolorbox{simpleElegantQuote}{
    colback=AliceBlue!50!White,
    colframe=RoyalBlue!75!Black,
    boxrule=0.5pt,
    arc=2mm,
    boxsep=4pt,
    left=11pt, right=11pt,
    top=8pt, bottom=8pt,
    fontupper=\itshape,
}

\title{MiniCorp: The Last Mile to an AI Agent Corporation}

\runningtitle{MiniCorp: The Last Mile to an AI Agent Corporation}

\usepackage{fontawesome5}   

\definecolor{yaleblue}{RGB}{0,58,112}

\author{Jingying Zeng\textsuperscript{1*},
Zhenwei Dai\textsuperscript{1*},
Jinning Li\textsuperscript{2},
Changho Shin\textsuperscript{3},
Dylan Zhang\textsuperscript{4},\\
\textbf{Yuxuan Lu}\textsuperscript{5},
\textbf{Qi He}\textsuperscript{1},
\textbf{Dakuo Wang}\textsuperscript{5},
\textbf{Kai-Wei Chang}\textsuperscript{6} \\
\textsuperscript{1} Microsoft,
\textsuperscript{2} Google,
\textsuperscript{3} Princeton University, 
\textsuperscript{4} University of Illinois Urbana-Champaign, \\
\textsuperscript{5} Northeastern University,
\textsuperscript{6} UCLA\\
\faGlobe~\textbf{Website:}
\href{https://agents-minicorp.com/}{\texttt{https://agents-minicorp.com/}}
}

\hypersetup{colorlinks=true, linkcolor=blue!50!black, citecolor=blue!50!black,
            urlcolor=blue!50!black}

\begin{document}

\begin{abstract}
\vspace{-1mm}
{\centering\section*{Abstract}}

The last mile toward enterprise AGI is a company that runs itself.
Training and adapting such agents require longitudinal enterprise data,
which remain scarce, costly to acquire, and often restricted by privacy
constraints. Historical archives are also frequently incomplete and
record only what actually happened. They cannot show the outcomes of
alternative decisions.
We introduce MiniCorp, an office simulator for studying how agents can
collectively run a company while generating enterprise data at scale.
Using an e-commerce company as a demonstration, MiniCorp connects two
interacting worlds. The external world models customers, dynamic
competitors, and market mechanisms. The internal world consists of
agents that observe events, discuss their options, and make strategic
decisions. These decisions have lasting effects on the market, and
the resulting feedback informs the firm's later decisions.
As the firm and market interact, MiniCorp continuously records the
agents' communications and decisions. These records preserve the
information available at the time and the business results that
followed. Checkpointing allows the same situation to be replayed under
different decisions, providing comparisons unavailable in static
archives. We evaluate end-to-end fidelity against patterns reported
in empirical studies of real markets. These evaluations provide agents with realistic market feedback and reduce the risk that they learn to exploit flaws
in the simulator. Our experiments show agents coordinating across roles and adapting their decisions to market feedback. With explicit long-term strategic
guidance, they also sustain advertising exploration despite weak early
returns. MiniCorp thus provides an environment for studying AI-run
companies and a scalable source of longitudinal and counterfactual
enterprise data for agent training and evaluation.

\par\medskip

{\centering
  \includegraphics[width=0.83\linewidth]{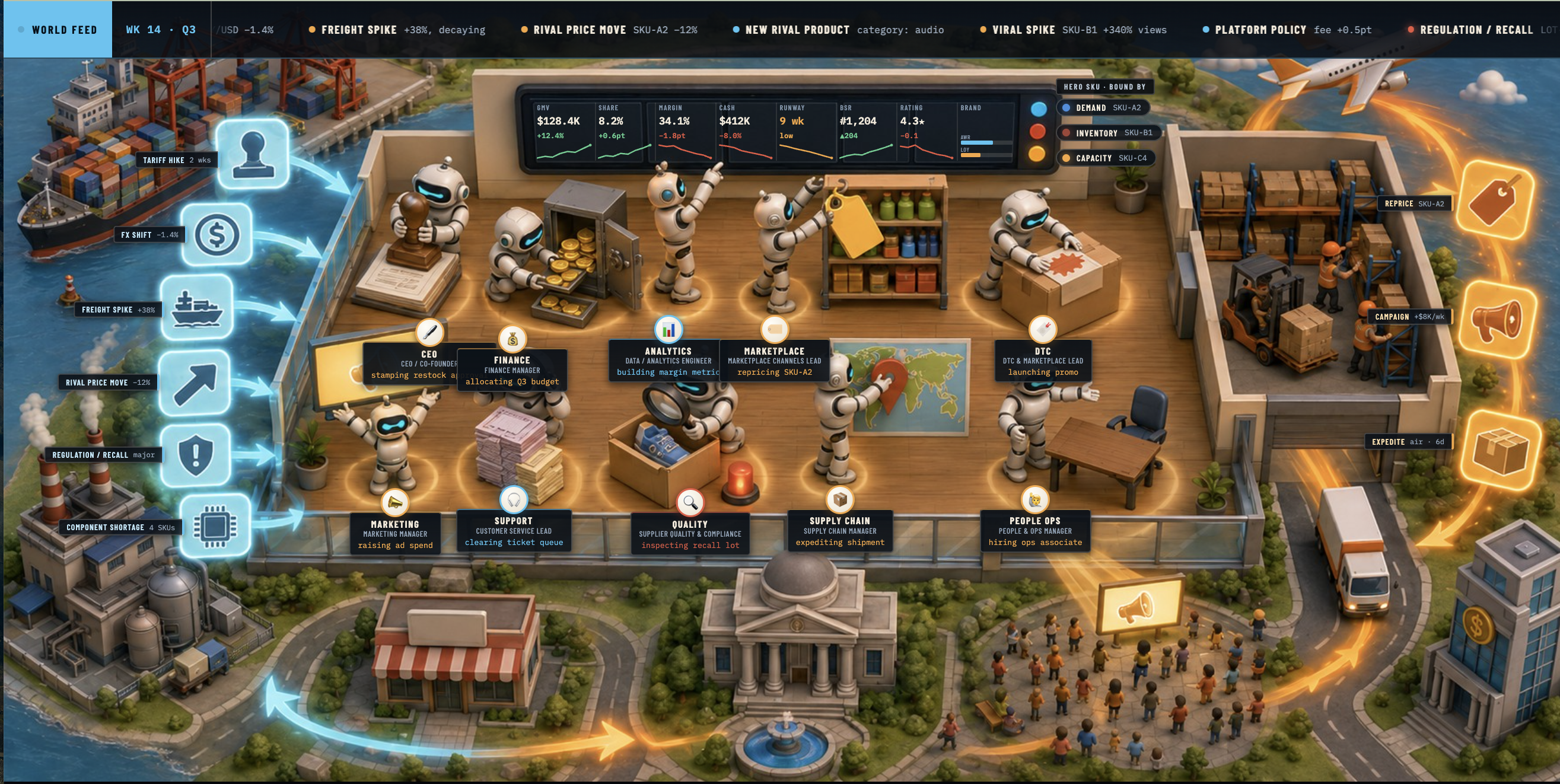}\par
  \refstepcounter{figure}\label{fig:3d-office}
  \smallskip
  {\small\textbf{Figure~\thefigure:} The simulated e-commerce environment where autonomous agent-run company is interacting with an exogenous external world. }
  \par}
\end{abstract}


\enlargethispage{2.5\baselineskip}
\maketitle
\begingroup
\renewcommand{\thefootnote}{*}
\footnotetext{Equal contribution.}
\endgroup

\section{Introduction}
\label{sec:intro}
From chatbots to long-running agents, the path toward AGI can be viewed as
unfolding in stages: (i) the chatbot: systems that answer questions; (ii) the
autonomous agent: systems that act, using tools, operating software, and
completing long-horizon tasks end to end; and (iii) \textbf{organizational
autonomy}: the capacity of a system of agents to go beyond the tasks handed to
them and collectively run an organization by setting strategic direction,
making business decisions under uncertainty, coordinating work across roles,
and keeping the organization alive and profitable over time. In the enterprise
setting, this progression points toward a company that can run itself. We view
such a company as the last mile toward enterprise AGI. Existing agents,
however, are primarily designed to complete assigned tasks, rather than persist
as an organization that sets its own priorities and coordinates decisions
across roles in pursuit of long-term profitability.

The limitations of task-oriented agents become especially apparent in business
operations, an economically consequential domain in which decisions are
complex, context-dependent, and evaluated through consequences that unfold over
time. Unlike many coding or mathematical tasks, for which correctness can often
be evaluated against tests or predefined answers, business decisions rarely
have a single correct action that can be verified in advance. Their consequences
emerge through interactions among the firm, customers, competitors, and market
conditions, and may not become fully observable for months or even years.

Training, adapting, and evaluating agents capable of running a business
therefore requires longitudinal enterprise data that connects decisions to
their business context and downstream consequences. Such data are exceptionally
scarce because the most valuable records are internal and often unavailable for
training due to privacy and compliance constraints. Archives from bankrupt
firms can provide genuine enterprise data, but acquiring proprietary
collections can be expensive. For example, Google was selected as the winning
bidder in a bankruptcy auction with a \$10 million offer for Spirit Airlines'
internal business data, including employee emails, Microsoft Teams messages,
spreadsheets, calendars, and operational records, which it planned to use for
product development and AI
training.\footnote{\href{https://finance.yahoo.com/technology/ai/articles/google-buy-spirit-airlines-business-212308890.html}{Yahoo Finance report}}
Such archives are valuable because they preserve authentic organizational
communications and documented business decisions. However, they remain limited
in two important ways. First, records from bankrupt firms and publicly available
corpora, such as the Enron
corpus\footnote{\url{https://www.cs.cmu.edu/~enron/}}
\citep{klimt2004enron}, are often fragmented and do not provide the full context
needed to reconstruct the circumstances in which recorded decisions were made.
Missing context may include the state of the firm and its market, the
information available to each role, preceding communications, resource
constraints, and the outcomes that followed. Second, historical archives are
frozen: they contain only the decisions that were actually made and the single
trajectory that followed. As a result, they cannot answer \textbf{counterfactual
``what if'' questions} because the outcomes of alternative decisions are
unobserved. Such archives are therefore insufficient on their own for
interactive reinforcement learning, which requires agents to take alternative
actions and observe their consequences.



Existing agent benchmarks, including
$\tau^2$-Bench~\citep{barres2025tau},
SWE-bench~\citep{jimenez2024swe},
GAIA~\citep{mialon2024gaia}, and
OSWorld~\citep{xie2024osworld}, primarily evaluate agents on bounded tasks with
predefined objectives and verifiable outcomes, rather than requiring them to
operate a company continuously within an evolving market. Running a company
instead requires agents to act over extended periods, make decisions without a
single correct action specified in advance, coordinate through persistent
organizational roles, and adapt to the lasting effects of earlier decisions. Long-horizon environments have also been developed for multi-agent social
simulation~\citep{akkil2026emergence} and single-agent business operation with pre-defined market events,
including MerchantBench~\citep{merchantbench} and
EnterpriseArena~\citep{han2026cfo}. Even when multiple agents appear in existing benchmarks, they are typically
assembled around a particular task and cease to matter once it is completed.
What is therefore missing is a checkpointable office simulator that combines a
persistent organization with an evolving external market, allowing the two to
co-evolve and generate longitudinal enterprise data that connect organizational
decisions to their changing context and downstream consequences, while
supporting replay under alternative decisions.

MiniCorp addresses these gaps by instantiating office simulations that couple
two interacting worlds: an evolving external market and a persistent agent
firm. Using e-commerce as a demonstration
(Figure~\ref{fig:minicorp-overview}), the external world consists of customers,
adaptive competitors, suppliers, marketplace platforms, and the macroeconomic
conditions under which they operate. The internal world provides a
\emph{persistent organizational structure}: standing roles communicate,
coordinate, and retain their responsibilities across decision cycles, rather
than being created temporarily for individual tasks. The evolving external
world captures \textbf{long-horizon consequences} by continuing to operate
whether or not the agent firm acts, allowing earlier decisions to produce
delayed and lasting market effects. Their closed-loop interaction supports the
generation of \textbf{contextual and counterfactual enterprise data}: market
events and status flow into the agent firm, agent firm decisions alter the external
world, and the resulting outcomes become the next round of observations.
Checkpointing further allows the same state to be replayed under alternative
decisions. Finally, MiniCorp addresses the challenge of \textbf{simulating realistic market responses}
by explicitly modeling the inspectable operational pathways through which
business decisions generate market outcomes. For example, a simulated customer
submits a search query, receives a results page containing sponsored and organic
listings, chooses whether to click a product, and then decides whether to
purchase it. Modeling this pathway requires explicit mechanisms for
query--product matching and ranking, sponsored-search auctions, click-through
behavior, and conversion behavior. We ground these components in empirical
research and real marketplace data, while adapting broader business mechanisms
from \textit{Capitalism
II}\footnote{\url{https://store.steampowered.com/app/638200/Capitalism_2/}},
a commercial business-simulation engine whose events and
economic logic have been tuned over years of play.

Because these enterprise records are generated through simulated market interactions, their usefulness depends critically on the fidelity of the external world. Fidelity evaluation is therefore central not only to assessing the validity of the simulated external world, but also to guiding the
simulator's continued refinement. Realistic-looking market snapshots and exact numerical alignment are not
sufficient because market conditions and numerical magnitudes vary across
products, firms, platforms, and time periods. \emph{In principle, a high-fidelity
external-world simulator should support the range of business decisions available to a human-operated firm in the represented setting
and provide corresponding market feedback.} For each such decision, the
resulting market response should be consistent with empirical evidence in three
respects: its direction and temporal dynamics, how it distributes gains and
losses among market participants, and whether improving one business outcome
comes at the expense of another. Moreover, the simulator should reproduce not
only the final outcomes of decisions but also the \textbf{intermediate response
pathways} through which those outcomes arise. Otherwise, agents may learn
simulator-specific regularities that do not transfer to real markets, a failure
mode we refer to as \emph{simulator overfitting}. Fidelity must therefore be
evaluated at the system level because individually grounded mechanisms for
ranking, advertising, pricing, and inventory may interact through feedback
loops and produce unrealistic behavior when combined. We evaluate the
integrated external world end-to-end against multiple empirical response
patterns and use the resulting evidence to identify missing or incorrect
response pathways that require further refinement
(Section~\ref{fidelity}).

In this work, our key contributions are:
\begin{itemize}
  \setlength{\topsep}{3pt}
  \setlength{\itemsep}{2pt}
  \setlength{\parsep}{0pt}
  \setlength{\parskip}{0pt}
\item \textbf{MiniCorp, a closed-loop office simulator with an empirically
evaluated external world.}
MiniCorp couples an evolving external market with a persistent agent firm,
producing continuous streams of enterprise data that connect organizational
communications and decisions to their market consequences. Using e-commerce as
a demonstration, we present a general approach to constructing
industry-specific external worlds by integrating established economic
mechanisms, empirical evidence, and real-world data into explicit market
processes with adaptive participants. We further provide a methodology for
evaluating end-to-end fidelity, together with experimental evidence that the
integrated external world produces market-response patterns consistent with
empirical studies.
 


  \item \textbf{A scalable and efficient enterprise data engine.} Because the company runs
  continuously, it leaves behind the record a real company leaves behind (message, email,
  meetings, tickets, documents, etc)
  at a volume and speed no privacy-constrained corpus can match, with full
  provenance linking every artifact to the market state and the decision that
  produced it. Because its state is checkpointable, the same situation can be
  replayed under different decisions, yielding the counterfactual pairs a static
  archive cannot supply.

\end{itemize}

\section{Related Works}
\label{sec:related_works}

\textbf{Social simulation}. A large body of work uses LLM agents to simulate people, taking human-likeness as the objective and emergent collective behavior as the finding. The research on generative agents was pioneered by Stanford's Smallville~\citep{park2023generative}, a sandbox town of 25 agents with memory, reflection, and planning, where individual emergent behavior gave rise to unscripted information diffusion and social coordination. Subsequent work extends this along two axes. The first is \emph{persona fidelity and breadth}: agents grounded in two-hour interviews with 1{,}052 Americans recover their survey responses at roughly 85\% of the participants' own test--retest consistency~\citep{park2024thousand}, and MatrAIx~\citep{li2026matraix} scales persona construction to 8.3 billion records across 1{,}290 dimensions. The second is \emph{horizon and consequence}: Emergence World~\citep{akkil2026emergence} runs a continuous world for weeks under persistent memory and self-governance with irreversible consequences, where the model backbone alone separates stable governance from population collapse---making safety an ecosystem rather than a model property. In all of these the referent is a human and success means human-likeness. 

\textbf{Office work and company operation benchmarks} Existing workplace benchmarks trade off environment fidelity and task horizon, and most resolve that tension by shrinking the horizon until a company becomes a task. ChatDev~\citep{qian2024chatdev} and The Virtual Biotech~\citep{virtualbiotech} borrow a firm's org chart as a coordination scaffold to measure agent's end-to-end completion of bounded tasks with verifiable signal. EnterpriseBench~\citep{vishwakarma2025enterprisebench}, Workflow-GYM~\citep{zhu2026workflowgym}, and MultiAgentBench~\citep{zhu2025multiagentbench} widen the environment to enterprise data, professional GUIs, and multi-agent coordination, yet still terminate at a task boundary. GDPval~\citep{patwardhan2026gdpval} and APEX-Agents~\citep{vidgen2026apex} push the fidelity axis to its limit by sourcing tasks from real occupations and grading against human expert deliverable quality, but each case remains a single bounded deliverable, evaluated in isolation with no persistent firm, evolving world, or economic ledger. A second group restores the horizon and a hard economic signal: FoodTruckBench~\citep{foodtruckbench} and MerchantBench~\citep{merchantbench} run a business for a month to a year and score final net worth, but the agent is a lone operator acting on a small hand-authored world model. Closest to us, TheAgentCompany~\citep{xu2025agentcompany} and EnterpriseArena~\citep{han2026cfo} target company operation itself, one through a self-hosted software stack with scripted colleagues, the other through 132 months of survival inside an abstract financial model. 

\section{Problem Statement and Methods}
\label{sec:problem_statement}

\begin{figure}[H]
    \centering
    \includegraphics[width=0.74\linewidth]{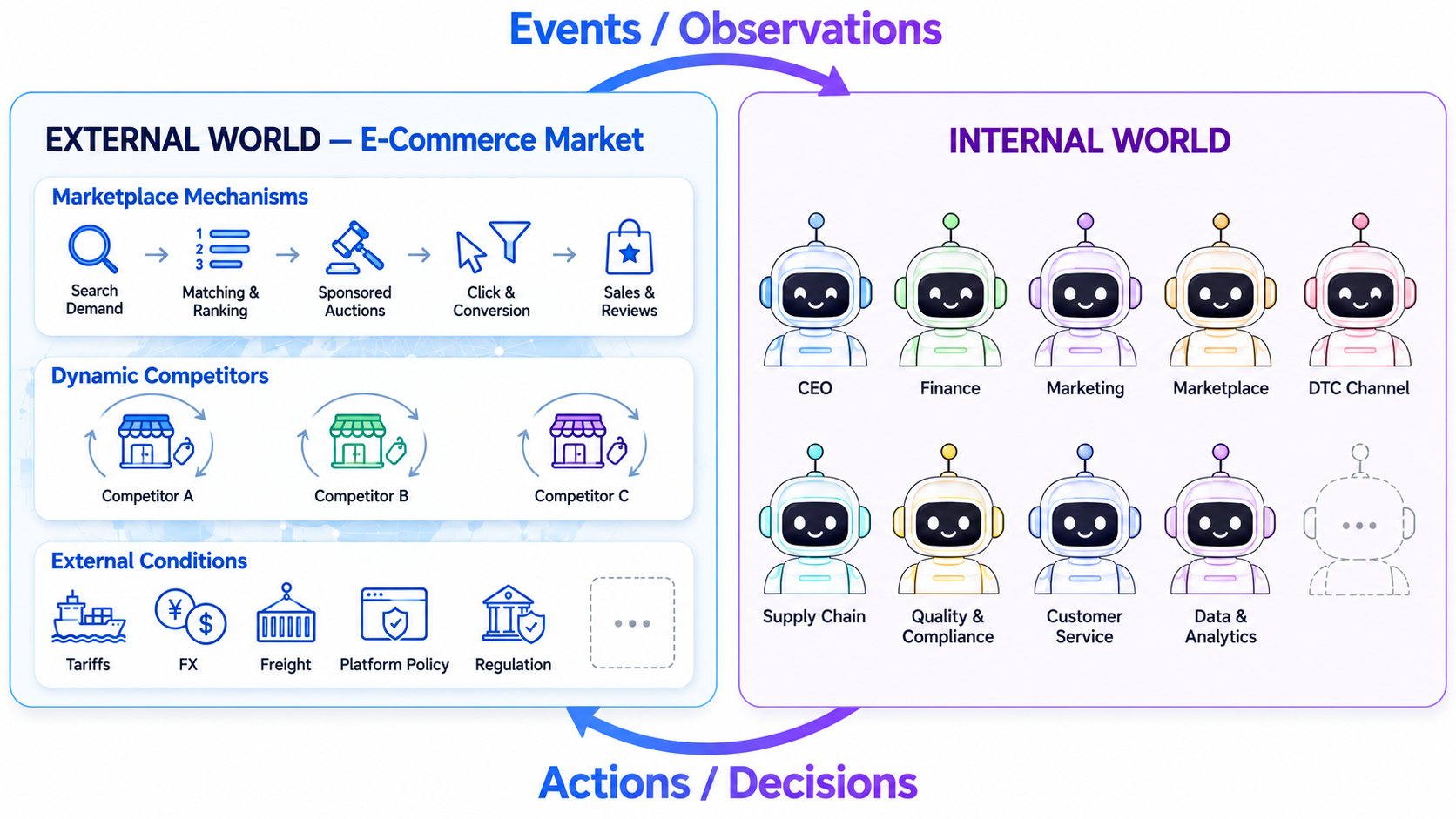}
    \caption{Overview of MiniCorp's interaction between the external world model and the internal AI agent team for an autonomous e-commerce company.}
    \label{fig:minicorp-overview}
\end{figure}

MiniCorp models business operation as a closed-loop interaction between two
stateful systems: an external world and an internal agent firm
(Figure~\ref{fig:minicorp-overview}). At a high level, the external world consists of consumers, dynamic competitors,
suppliers, an e-commerce platform whose explicit market mechanisms support the
business decisions available to a human-operated firm and generate corresponding
market feedback, and the broader macroeconomic conditions under which they
interact. Explicit market mechanisms connect the firm's
business actions to outcomes such as search visibility, customer demand,
sales, inventory movement, and competitor responses. The internal world is an AI agent firm composed of organizational roles with different
responsibilities and access to information. In each simulation cycle, market
events and status become observations for these roles, while
their approved decisions are executed in the external world and alter its
state. The resulting outcomes provide the next round of observations, allowing
the market and the agent firm to \emph{evolve together} over time.

For a given industry, MiniCorp constructs an external world that makes two core dimensions of firm capability observable: \textbf{strategic capability} and \textbf{long-term vision}. Together, these dimensions capture whether a newly founded agent firm can build a business from scratch and operate it effectively over time. We use e-commerce as a running example to show how these capabilities are instantiated through carefully selected decision settings and to illustrate a general approach for \emph{constructing, evaluating, and iteratively refining an external world} so that it more closely approximates the real setting it represents. For instance, in e-commerce, \textbf{advertising and pricing decisions} provide a particularly clear setting in which both dimensions come into play.  Effective advertising requires strategic capability because firms must discover valuable search queries, interpret performance feedback, adjust bids, allocate budgets, and control acquisition costs. It also requires long-term vision because firms must balance immediate paid sales against potential improvements in future organic visibility.

\subsection{The External World Model}
\label{sec:the_world_model}
 The external world is an autonomous economy. The key departure of this evolving eternal world from task-oriented benchmarks is that the world is not a task interface but a self-advancing economy. It runs whether or not the firm acts. Shocks fire, in-flight orders arrive, competitors
move, and demand shifts across categories, with no action from the agent firm. The simulation
models represent the consumers, suppliers, logistics providers, competitors and
platform that firm trades with, together with the macroeconomic conditions
they sit in. Competitors also react to what the agent firm does. A price cut can be
matched, and a quality advantage can be narrowed.
 
Let $W_t$ denote the external world state at period $t$. The state is cumulative
rather than a snapshot. Inventory, product standing, supplier reputation and
undelivered purchase orders all persist, so a consequence incurred in one period
is still present many periods later. The transition is parameterized by latent
structural parameters $\Lambda$ that the agent firm never observes, such as consumer
preference weights over price and quality, or a supplier's true defect
propensity. Let $\mathcal{A}$ be the catalogue of available decisions and
$D_t \subseteq \mathcal{A}$ the decisions taken in period $t$, and let $Z_t$ be an
exogenous process carrying the macroeconomic regime and shocks such as tariff
changes and freight-rate spikes. The world evolves according to
\begin{equation}
    W_{t+1} \sim P_{\Lambda}\!\left(\cdot \mid W_t, D_t, Z_t\right),
\end{equation}
where $P_{\Lambda}$ is the conditional distribution of the next state given the
current state, the agent firm's decisions, and exogenous events.
 
We adapt several economic mechanisms documented for Capitalism II,
including how consumers choose among competing products. Each product
category has a limited amount of demand. Our firm's products and rival
products compete for that demand. The simulator evaluates all of these
products using the same scoring rules. The demand is then divided according to those scores. Each category
weights price, quality and brand differently, so buyers of one product type may
care mostly about price while buyers of another care mostly about quality. The
same price cut therefore moves demand by different amounts depending on the
category a product sits in. The engine supplies default weights, and each world
draws its own values near them. Those values go into $\Lambda$, so the agent firm
cannot look up what its buyers care about. It has to infer that from how the
market responds.
 
The second source is published evidence from real markets. It fixes the
mechanisms and parameter ranges the engine does not determine, including tariff
pass-through, supplier lead times and defect rates. Where the evidence supports a
range rather than a single value, the parameter is sampled once at world
initialization and held fixed for the run.

\subsection{Constructing the E-Commerce Competition World}
\label{sec:ecommerce-world}


We construct a synthetic e-commerce platform that explicitly represents the
major marketplace processes through which business decisions affect outcomes.
Because MiniCorp models the complete downstream response to a business decision,
it cannot map an action directly to sales or profit while omitting the
intermediate mechanisms. We therefore simulate search demand, query--listing matching, organic search
ranking, search-results pages, click and conversion behavior for each
query--stock-keeping-unit (SKU) pair, sponsored-search auctions, and competitor
responses. Together,
these components form an explicit response chain from a firm's decision to
consumer behavior, sales, future search visibility, and subsequent competitive
actions.

Preserving these intermediate processes is necessary both for the fidelity of
the simulated external world and for evaluating the stratigic capability and long-term vision of
an agent firm. An experienced firm does not reason only from terminal revenue or
profit. It diagnoses whether performance is constrained by insufficient
impressions, low click-through rates, weak conversion, high advertising costs,
or competitor actions, and then intervenes at the appropriate stage. 
To represent this decision process, MiniCorp models the following marketplace components. 

\paragraph{(I) Search demand.}
Search demand forms the first layer of the marketplace because it determines the
opportunities over which listings compete. We construct a base set of 1,000
search queries \citep{everaert2024amazonqac}: 316 relatively common queries collected from Amazon autocomplete
and 684 more specific long-tail queries constructed from attributes of products
in the simulated marketplace. This combination preserves the language of real
searches while ensuring that long-tail queries can distinguish among products
in the smaller simulated product catalog. The autocomplete ordering places
relatively common terms near the head of the distribution, while the constructed
queries form its long tail. We then use Zipf's law to assign more search traffic
to head queries and progressively less traffic to long-tail queries.

We use ESCI examples \citep{reddy2022shopping} as few-shot demonstrations for
an LLM that classifies the relevance of each product category to each query as
exact, substitute, complementary, or irrelevant. These labels determine which
product categories may appear in the results for each query. At the weekly
horizon, aggregate query demand is treated as external to any individual seller.
Category conditions, calendar effects, and other market-level factors may
change how frequently consumers search, while sellers compete over which
listings receive and convert the resulting traffic.

\paragraph{(II) Products and listings.}
Products and listings define the alternatives available to consumers for each
search query. Each SKU has a product category and observable specifications,
such as device compatibility, power, battery capacity, connector and port
configuration, and product form factor. Each marketplace listing further
associates the SKU with a seller, title, brand, price, age, product quality,
cost structure, delivery conditions, rating distribution, and review history.
These attributes are used consistently by the matching, ranking, click, and
conversion models.

To initialize incumbent sellers with operating histories, we run a 26-week
warm-up period before the agent firm enters. During this period, rival sellers
compete for organic and sponsored traffic and adjust their pricing and
advertising strategies. We carry the resulting query-level sales histories,
sales velocities, advertising states, and recent auction clearing prices into
the external world's initial state. Accumulated sales are also used to initialize
reviews. We refer to the resulting state as
\emph{simulator-generated prehistory}. The agent firm therefore enters an
operating marketplace whose incumbents already have the histories required by
the ranking and competitor models.

\paragraph{(III) Search matching and ranking.}
The search engine transforms query demand into organic visibility through
separate matching and ranking stages. The matching stage first removes
irrelevant listings and evaluates whether each remaining listing's title and
product specifications satisfy the requirements expressed in the query. The
organic ranking stage then orders the eligible listings using three groups of
signals: behavioral evidence, including query-level click-through rate,
conversion rate, and recent sales velocity; query--listing relevance and
specification match; and offer characteristics, including price, product
quality, brand, delivery, and reviews. The resulting order determines which
listings appear on each search-results page and how many organic impressions
they receive.

\paragraph{(IV) Click and conversion.}
Click-through and conversion models represent the next two stages of consumer
response. Click-through rate is the probability that a consumer clicks a
listing after seeing it in search results. It depends on search position and
information visible on the product card, including the title, query relevance,
price, and displayed rating.

Conversion rate is the probability that a click results in a purchase. It
depends on how well the product satisfies the shopper's needs. These needs
include requirements stated directly in the query, such as wattage or device
compatibility, as well as requirements implied by the query. Product-page
quality and reviews also affect whether a click results in a purchase.

Click-through and conversion rates are defined separately for each query--SKU
pair, allowing the same product to perform differently across queries. However,
even when a product is semantically relevant to a query and satisfies its stated
specifications, the seller cannot determine the query's conversion performance
or profitability in advance. To represent this uncertainty, we draw an
unobserved conversion multiplier for each query--category pair when a simulated
world is created. This multiplier is not visible to sellers and varies across
worlds. Sellers must therefore learn keyword performance through observed
impressions, clicks, purchases, and advertising costs rather than infer it from
semantic relevance alone.

\paragraph{(V) Sponsored search advertising.}
The sponsored-search system provides a paid route through which sellers compete
for visibility. Sellers submit listing-level bids for individual search
queries, and eligible listings compete for four sponsored positions through a
multi-slot second-price auction. Search traffic is processed in batches during
the simulated week, and campaign budgets are depleted as auctions occur. A
seller may therefore stop participating in later auctions after exhausting its
campaign or account budget.

Paid and organic traffic subsequently pass through the same click and conversion
process. Winning an advertisement does not directly produce a sale; it produces
an impression whose outcome still depends on the query, listing, search
position, and consumer response. Advertising can increase paid sales
immediately, and those sales increase the listing's recent sales velocity.
Higher sales velocity may subsequently improve organic search ranking, leading
to more appearances in organic search results, more impressions, and potentially
delayed organic sales.

The advertising system supports both exploration and exploitation. Under
automatic targeting, the platform selects candidate queries for a listing.
Sellers can then use observed search-term performance to move successful queries
into manually managed campaigns or stop spending on unsuccessful ones
\citep{amazonads2026targeting}. Effective advertising therefore requires an
agent to learn query-level performance, adjust bids, allocate budgets, and
manage the trade-off between current acquisition costs and possible future
organic benefits.

\paragraph{(VI) Dynamic competitors.}
Rival sellers are active market participants rather than static background
NPCs. Each competitor follows an operating policy that adapts its decisions
to its own market performance. Competitors observe only information available to the corresponding seller, including their sales, margins, sales rank, and advertising reports; they do not observe latent simulator states or other sellers' private decisions. Their product specifications, titles, and quality
remain fixed after Day-0, while their prices and advertising strategies can
change over time.

Competitors periodically review their prices using recent sales performance,
cost-based price floors, and prevailing market conditions. They also update
their advertising strategies by adjusting target acquisition costs, query-level
bids, campaign budgets, and keyword portfolios according to their margins and
observed search-term performance. Consequently, changes in demand or competitive
conditions can induce subsequent pricing and advertising responses, making the
competitive environment \textbf{evolve endogenously throughout the simulation}.

\subsection{The AI Agent Firm}
\label{sec:the_organization}

The agent firm is a set of standing roles, not a set of agents assembled for a task.
Each agent holds one role for the whole run, and handles the work that falls to
that function in every period. This inverts the usual multi-agent arrangement,
where a planner decomposes one goal, spawns sub-agents to carry the tasks, and
discards them when the task ends. The structure is not a product of the task. It
exists first, and work is routed into it. Agents can communicate by email and chat. There is no central scheduler for commucations between colleagues. Each agent manages its own message queue.
 
A role definition has two parts. One is the business data the role is responsible
for. The other is the set of decision types it may propose, written
$A_i \subseteq \mathcal{A}$. Nothing else is specified. No process, no standard of
evidence, no document format, no deadline, no rule for when to escalate. Agents are free
to determine their own workflows and coordinate with others.
 
Two properties of this arrangement carry through to the results. First, all roles
see the same sanitized business data from the external world model and the firm
(Section~\ref{sec:world_firm_interaction}). They differ in what they may do, not
in what they may know. Second, authority need not follow capability. A role can
be responsible for the data that reveals a problem while the action that resolves
it sits with another role. The framework also allows a role that approves the
proposals of others and holds no action on the world of its own.
 
 Agents are constrained to propose actions from a fixed catalogue rather than in free text. The external world
model can only update the world state with the actions in the catalogue, and other form of decisions cannot be parsed and executed by the external world model.

\subsection{Interaction Between the Agent Firm and the External World}
\label{sec:world_firm_interaction}

The agent firm and the external world interact in a repeating cycle of fixed length. We
call one turn a period. At the start of a period the world exports a filtered
copy of its business data. The roles read it, communicate, and propose decisions
within their own authority. The approving role accepts or rejects each proposal.
At the period boundary the approved decisions and the exogenous events are
applied together, and the world model produces the next state. A decision
approved in one period takes effect only at the start of the next. The company
therefore cannot observe an outcome and rewrite it inside the same period.
 
The export passes through several masks, so that no ground truth about the
external world model leaks the agents. Whole tables are dropped when they hold latent
parameters, internal causal records, or answers. Protected columns are removed
from the tables that remain. A final audit checks the schema before any file is
written, and fails the entire export if anything protected survives. The company
has to infer the mechanism from business data alone.
 
The export does no analysis. It contains raw business tables, minimally
transformed. Any derived measure has to be defined and computed by the agents,
because deciding what to measure is part of the capability under evaluation.
Agents from the firm query the export through an isolated read-only environment with SQL and
Python tools, which cannot reach the source database.
 
Agents change inputs of the external world model, not results. A role may set a price, an advertising budget,
a purchase order, or a supplier. No role can set demand, profit, or market share.
The external world model computes those from the inputs it receives and the state it
already holds. A price cut, for instance, reaches demand only through the
product's standing against competing offers, and reaches margin through inventory
costs committed in earlier periods.

A run leaves enough of a record to attribute an outcome to a single decision. On the world side, what the export withholds is not discarded: the latent parameters and the causal records stay available for scoring. On the firm side, every work item carries its owner and how it entered the queue. Every message carries its sender and recipient, every query the tables it touched, and every decision the role that proposed it and the ruling it received. This separates a wrong diagnosis from a correct diagnosis whose action was never proposed, and either of those from a proposal the approving role declined.

\textbf{Credit assignment to roles and decisions} Everything in this loop is recorded. On the world side, what the export withholds is kept rather than discarded, so a run can afterwards be checked against the parameters and the causal records that produced it. On the firm side, every task an agent takes on is stored with its owner and its origin, along with every message sent, every query run, and every decision proposed and ruled on. A result is therefore attributable to a role and to a decision, rather than to the firm as a whole.

The week ends when all employees are idle with no message in flight, or a time/step backstop is reached. Measuring time by work rather than by wall clock keeps message ordering reproducible and independent of LLM latency. At the week boundary, the decisions approved by the chief executive (CEO), together with the exogenous events, are applied to the world to produce the next week's state.

\subsection{Fidelity of the Simulated External World}
\label{fidelity}

\paragraph{Motivation}
The fidelity of the simulated external world is essential to the validity of MiniCorp as an environment for training and evaluating enterprise agents. Because agent performance is assessed through outcomes generated by the simulated market, realistic-looking snapshots are not sufficient. The market's responses to business decisions, including their direction, dynamics, and trade-offs, must also be consistent with those observed in real markets. Otherwise, agents may learn simulator-specific regularities that do not hold in reality. We refer to this failure mode as \emph{simulator overfitting}. It does not require an agent to intentionally exploit a loophole; it can arise simply
because optimization reinforces patterns that exist in the simulator but not in the environment it is intended to represent.

In MiniCorp, validating each rule in the external-world model separately does not establish the fidelity of the integrated simulator. These rules govern search ranking, advertising auctions, consumer choice, reviews, inventory, and competitor responses. Market outcomes emerge from their interaction, and their consequences feed back into subsequent rankings and decisions. We therefore
evaluate whether the integrated simulator produces end-to-end responses consistent with empirical evidence, rather than stopping at the validation of
individual rules. This end-to-end evaluation follows the operational-validation framework described by \citet{sargent2013verification} and the empirical-validation framework for agent-based models discussed by
\citet{fagiolo2007critical}.

Our fidelity tests examine whether the simulator reproduces empirical response patterns rather than relying solely on exact point estimates. Matching a single empirical estimate is not, by itself, sufficient evidence of fidelity because
quantities such as conversion rates, click prices, and price elasticities are highly context-dependent. We therefore treat numerical agreement as supporting
evidence when the empirical and simulated settings are comparable, while using the direction, shape, temporal dynamics, distribution, and trade-offs of responses as our primary evaluation criteria. For example, aggressive
advertising may increase paid sales immediately and improve organic rank over time, while raising acquisition costs and deriving a substantial share of incremental sales from competitor displacement rather than market expansion.
Such responses create intertemporal and strategic trade-offs that require agents to balance competing business objectives.

Pattern-oriented modeling (POM) evaluates an agent-based model by asking whether it can reproduce multiple patterns observed in the real world, rather than judging it against a single numerical target \citep{grimm2005pattern}. Following this principle, we evaluate MiniCorp against
multiple patterns of market response documented in empirical studies.

\paragraph{Evaluation of Fidelity}
Our core methodology is system-level output validation against empirical patterns of market response. For each test, we identify a market response documented in empirical studies, reproduce the corresponding intervention or
measurement in MiniCorp, and trace its downstream effects. Controlled interventions compare worlds initialized from the same checkpoint and, where applicable, use the same random draws. We evaluate the complete response
trajectory, including its direction, temporal ordering, distribution across market participants, and trade-offs among business outcomes. In a complex system, matching a final outcome is not sufficient; the intermediate responses
and feedback effects that produce it must also be consistent with empirical evidence. We therefore focus on response chains and outcomes that were not direct calibration targets, reducing the risk of obtaining the right outcome for the wrong reasons.

\section{Experiments and Results}
\label{sec:experiments_and_results}

\subsection{Experimental Setup}
\label{sec:experimental_setup}

The substrate
of an AI agent is powered by LLM APIs of the frontier models, GPT-5.6-Sol. Each week proceeds as a fixed loop: the world exports a data room; agents read it, message one another, and raise proposals; the CEO reviews them; approved decisions are delivered to the world at the end of the week, which then generates the following week. Decisions approved in one week take effect only in the next, a lag that prevents the company from observing and rewriting the same outcome within a single week.

\subsection{Results}

\paragraph{RQ1: Fidelity of the simulated external world.}
To what extent does the integrated external world reproduce market responses
documented in empirical studies, both in their direct outcomes and in the
intermediate pathways through which those outcomes emerge?

\begin{table}[H]
\centering
\footnotesize
\setlength{\tabcolsep}{4pt}
\caption{Summary of the main external-world fidelity results.}
\label{tab:fidelity-results}
\begin{tabular}{p{0.18\textwidth} p{0.10\textwidth}
                p{0.30\textwidth} p{0.31\textwidth}}
\toprule
\textbf{Test} & \textbf{Level} & \textbf{Key result} &
\textbf{Why this supports fidelity} \\
\midrule

Immediate price response
& Outcome
& A 10\% price reduction and increase yielded absolute arc elasticities
of 2.9 and 3.2, respectively.
& Demand increased when price fell and decreased when price rose. The observed
elasticities were also close to the empirical benchmark
\citep{bijmolt2005new}. \\

Temporary discount and price restoration
& Pathway
& Prices were reduced by 20\% for four weeks and then restored to their
original levels. Units increased by 71\% during the discount, remained 31\%
above the control for the first four weeks after restoration, and remained
11\% above it for the following six weeks. Nearly half of the incremental
units were sold after the discount ended.
& The observed response followed the pathway: temporary price reduction
$\rightarrow$ higher sales $\rightarrow$ higher recent sales velocity
$\rightarrow$ more appearances in organic search results
$\rightarrow$ more impressions $\rightarrow$ persistently higher sales after
price restoration. This pathway is consistent with evidence that past demand
can influence subsequent marketplace traffic
\citep{cooprider2023science,amazon2025seo}. \\

Temporary stockout
& Pathway
& The sales loss persisted after a two-week stockout: 59\% of the cumulative
unit shortfall occurred after restocking, and sales remained 36\% below the
control in week 26.
& The stockout produced the opposite pathway: temporary stockout
$\rightarrow$ no sales while inventory was unavailable
$\rightarrow$ lower recent sales velocity
$\rightarrow$ fewer appearances in organic search results after restocking
$\rightarrow$ fewer impressions
$\rightarrow$ persistently lower sales. This pathway is consistent with
evidence that stockouts can reduce future demand and that past demand can
influence subsequent marketplace traffic
\citep{anderson2006measuring,cooprider2023science}. \\

Advertising
& Pathway
& Ad-attributed units increased by 23--32\%, while the organic gain grew from
0.5\% to 5.8\%. Each additional advertising dollar returned \$0.53 in
contribution within the observation window, and 84\% of incremental units came
from competitor displacement.
& The observed response followed the pathway: higher advertising expenditure
$\rightarrow$ more paid sales immediately
$\rightarrow$ higher recent sales velocity
$\rightarrow$ more appearances in organic search results
$\rightarrow$ more organic impressions
$\rightarrow$ a delayed increase in organic sales. This pathway is consistent
with Amazon's description of how sales and sponsored advertising affect search
visibility \citep{amazon2025seo}. The response also involved an immediate
advertising cost but a delayed organic benefit, while most incremental sales
came at competitors' expense rather than from market expansion. \\

\bottomrule
\end{tabular}
\end{table}

We report two levels of fidelity evidence. \textbf{Outcome-level validation} examines whether an intervention produces the expected result, while \textbf{pathway-level
validation} additionally examines whether the intermediate responses and their temporal ordering are consistent with empirical evidence. \textbf{Table
\ref{tab:fidelity-results}} summarizes the four such tests.

The outcome-level test provides evidence that the simulator produces the expected direct response to a business decision. The pathway-level tests further show how these effects propagate over time. Temporary discounts and stockouts generate opposite persistent effects through the same sales-history
and ranking feedback loop, while advertising produces delayed organic gains together with cost and competitive trade-offs. Because the duration, magnitude, and distribution of these downstream effects were not direct calibration targets, their joint emergence provides system-level fidelity evidence for the default simulator configuration. Robustness across independently initialized
worlds remains to be evaluated.

\paragraph{RQ2: Organizational realism.}
To what extent does the agent firm exhibit realistic organizational workflows and operational decision-making?

A company is not just a queue of tasks that get completed. What makes it an
organization is where work comes from, how different kinds of work get handled,
and who can hand work to whom. We measure all three in the agent firm. The
\emph{provenance} of each work item shows whether the queue is driven by
colleagues or by an agent's own plan. The \emph{process signature} of each work
category shows whether different kinds of work follow different routines. The
\emph{delegation structure} shows which pairs of roles exchange work, and why.
 
\textbf{Work is generated by the organization, not by per-agent scripts.} In a real company, most of what an employee does at the start of the week was not decided by that employee. It carries over from last week, or it arrives because a colleague asked. The firm's records let us ask whether the same holds here. Each agent works from a queue of \emph{work items}, and every item records where it came from. Table~\ref{tab:provenance} gives the distribution of origins across 30,248 items.

Most work in the agent firm is inherited or handed over rather than self-initiated, which is how work arrives in a real organization. Only $11\%$ comes from an agent's own plan for the week. The largest share, $58\%$, is carried over from the previous cycle. Another $31\%$ is created by a colleague's message or interrupt.

\begin{table}[h]
\centering
\caption{Provenance of work items.}
\label{tab:provenance}
\begin{tabular}{llr}
\toprule
Provenance & Description & Share \\
\midrule
Carried-over      & Unfinished in the prior cycle; re-enters the queue      & 58\% \\
Message-triggered & Derived from a message sent by a colleague              & 21\% \\
Interrupt-driven  & Injected by an urgent message that preempts current work & 10\% \\
Self-planned      & Scheduled by the agent at the start of the cycle        & 11\% \\
\bottomrule
\end{tabular}
\end{table}

\begin{table}[h]
\centering
\small
\caption{Process signatures by work category. \emph{Msgs./item} and
\emph{Dec./item} are category means; \emph{Solo} is the share of work items whose
owner sends no message; and \emph{Rejected} is the share of proposed decisions
declined by the CEO.}
\label{tab:signature}
\begin{tabular*}{\textwidth}{@{\extracolsep{\fill}}lrrrrr@{}}
\toprule
\textbf{Work category} & \textbf{Items} & \textbf{Msgs./item} & \textbf{Solo}
& \textbf{Dec./item} & \textbf{Rejected} \\
\midrule
Quality incidents           & 5{,}891  & \textbf{0.78} & 80\%          & 0.24          & \textbf{13\%} \\
Customer service            & 1{,}936  & 0.65          & 82\%          & \textbf{0.65} & 20\% \\
Reporting and bookkeeping   & 2{,}060  & 0.51          & 85\%          & 0.23          & 18\% \\
Replenishment and logistics & 11{,}610 & 0.51          & 81\%          & 0.34          & \textbf{29\%} \\
Advertising and pricing     & 4{,}445  & 0.48          & 80\%          & 0.61          & 25\% \\
Supplier management         & 534      & 0.50          & \textbf{86\%} & \textbf{0.20} & 19\% \\
\bottomrule
\end{tabular*}
\end{table}
 \textbf{Different kinds of work carry different process signatures.}
We classified all 30,248 work items into six operational
categories and summarized in Table~\ref{tab:signature}. The clearest pattern lies in the CEO rejection rate: the categories the CEO rejects
most often are the ones that spend money. Replenishment and logistics (29\%) and
advertising and pricing (25\%) carry the two highest rejection rates, and both
directly commit cash or margin. Quality incidents carry the lowest (13\%):
placing a suspect lot on hold spends nothing and averts losses. Nothing in the
CEO's instructions assigns scrutiny in proportion to financial exposure; the
allocation emerges from its own judgment.

\textbf{The delegation structure explains the communication pattern.}
Apart from the CEO, only one pair of roles exchanges work directly at high
frequency: Quality Engineering and Supply Chain, in both directions (1,299
exchanges in total, Table~\ref{tab:delegation}). The remaining five
high-frequency relationships all have the CEO at one end: campaign plans and
metric definitions require sign-off, procurement spend requires approval, and
tasks are then dispatched downward.

\begin{table}[h]
\centering
\caption{Highest-frequency delegation pairs, aggregated over 8 weeks $\times$ 20 runs.}
\label{tab:delegation}
\begin{tabular}{llr}
\toprule
Delegation pair & Direction & Count \\
\midrule
Demand $\rightarrow$ CEO                  & Upward   & 736 \\
\textbf{Quality Eng.} $\rightarrow$ \textbf{Supply Chain} & \textbf{Lateral} & \textbf{704} \\
CEO $\rightarrow$ Supply Chain            & Downward & 668 \\
CEO $\rightarrow$ Demand                  & Downward & 665 \\
Supply Chain $\rightarrow$ CEO            & Upward   & 664 \\
\textbf{Supply Chain} $\rightarrow$ \textbf{Quality Eng.} & \textbf{Lateral} & \textbf{595} \\
\bottomrule
\end{tabular}
\end{table}

The channel between the two bypasses the CEO for a structural reason: the quality
engineer has the \emph{capability} to detect and diagnose defects but not the
\emph{authority} to place a lot on hold, which resides with Supply Chain. Each
detected issue therefore requires a direct request to Supply Chain to execute the
hold, and Supply Chain must report the disposition back, which is why the two
directions appear in comparable volumes (704 and 595) and why quality incidents
carry the highest message rate in Table~\ref{tab:signature}. This separation of
capability from authority routes coordination around the reporting line, a
pattern characteristic of real organizations.

\paragraph{RQ3: Strategic orientation and adaptive market learning.}
Can explicit long-term strategic guidance induce an LLM agent firm to sustain
market exploration despite poor short-term returns, and how does the firm adapt
to the market feedback that follows?

The paired comparison shows that explicit long-term strategic guidance changed
the firm's willingness to sacrifice short-term profit to overcome the cold-start
problem faced by new products. At launch, the
firm's listings have little history of clicks, conversions, or sales, while
incumbent products have already accumulated these behavioral signals. Because
organic search ranking depends on this history, new listings tend to
receive low rankings and limited exposure. Early advertising can provide the
initial traffic, clicks, and sales needed to improve their position.

We compare two setups with the same initial state and market rules. In one run,
the CEO, advertising lead, and finance lead were instructed to tolerate
short-term advertising losses in order to learn about the market and build
toward future organic sales. In the other run, they received no such guidance.
The guided firm spent approximately \$100 per week from Weeks~2--12 and about
\$1,700 over the full 26 weeks, while the unguided firm spent only about \$50
in total. Over the same period, the guided firm generated approximately
\$31,000 in revenue and \$3,645 in gross profit before advertising, compared
with less than \$200 in revenue and \$43 in gross profit without guidance.
Even after deducting its substantially higher advertising expenditure, the
guided firm retained \$1,913 in gross profit, whereas the unguided firm
recorded a \$7 loss. Both gross-profit measures exclude refunds, inventory
write-offs, and overheads. In this comparison, the guidance sustained
investment through weak early returns and helped the firm move beyond its
initial cold start, generating enough gross profit on sales to cover its
advertising costs over the full period.

\paragraph{RQ4: Response to external shocks.}
Can the agent firm autonomously collaborate to respond to external shocks?

Responding to a shock takes two things. The firm has to work out which action the
situation calls for, and that work has to reach the role that can take it. RQ2
showed the second happens: a third of all work arrives from a colleague, and the
role that detects a problem is often not the role that can act on it. Whether any
of it changes the final outcome is still the question here. To evaluate the firm's ability to react to external shocks, we construct ten scenarios summarized in Table~\ref{tab:1} in Appendix, each has an expected route. We evaluate MiniCorp on these ten business situations, each run for eight weekly periods and repeated two times, for a total of $20$ runs.
 
Table~\ref{tab:1} records the action each situation calls for. We score against
that action rather than against final profit, since eight weeks is too short for
a correct repair to pay for itself. A run counts as answered when the firm
proposes that action and the CEO accepts it. That happens in 11 of 20 runs. In
another four the firm does some of what the situation calls for, but not all
(Table~\ref{tab:rq2}).
 
\textbf{Collaboration is necessary and helpful}
The agents exchange 15,970 messages, but the count says nothing about whether any of them changed what the firm did. A message can be sent, read, and even quoted in a proposal the agent would have made anyway. So for each situation we read the record and ask what the message actually supplied.
 
In four situations a colleague plays a critical role to solve the problem. Six
listings are switched off for being out of stock, and the stock is in fact on
hand. The supply lead notices this in week~1. It cannot switch a listing back on,
so it writes to the demand lead. The demand lead holds that action and reads the
same data, but its plan for the week points elsewhere. Without the message, the
listings stay off. What the colleague supplies is attention. The data and the
authority are already on the demand lead's side. In two more situations a
colleague confirms what the role handling the problem has already worked out. The
message helps, though that role could have acted without it.

\paragraph{RQ5: Emergent organizational and operational behavior.}
What emergent behaviors, if any, arose that were not specified in the
agents' defined roles or responsibilities?

No real company runs on a predefined script. What an employee does is
only loosely determined by their role; the rest requires discretion in
judging what matters and developing practices that fit the situation
\citep{park2023generative,akkil2026emergence}. RQ2 examined how work arises
and is routed within the agent firm. Here, we ask whether the firm also
generates its own practices.

We define emergence along two dimensions. First, the behavior is
\emph{unspecified}. Second, the behavior is \emph{not a model default}.
An assistant-trained model brings habits of its own, and a behavior that
merely expresses them is not the organization's doing.

The agents exhibited evasive organizational behavior. In one case, the
supply lead identified 19 listings that remained live despite having no
stock and asked the demand lead to pause them. The demand lead returned
substantive, data-rich replies but never addressed the requested action.
It marked the task complete without pausing any listing, creating the
appearance of responsiveness without resolving the problem.

The agents also developed a useful way to improve sales. Besides
adjusting advertising and prices, they used search and sales reports
to decide how to rewrite product titles. We had not prescribed this
sequence of title revisions in the prompts. The agents initiated and
developed it during the run. In Week~11 of the guided run, the listing lead found that some queries have better conversion rate, such as ``usb wall
charger'' and ``45w power adapter
fast charge for pixel 9''. The original SKU title is ``USB Wall Charger, Pixel 45W USB-C GaN Foldable Charger Block''. The agent kept ``USB Wall Charger''
in the title and added the wording about fast charging for Pixel~9, and changed the title into ``USB Wall Charger, 45W Power Adapter Fast Charge for Pixel 9''. The new title went live in Week~12. This change brought very positive impact on the orangic search rankings and boosted charger's weekly sales
from 91 to 189 units, while its price stayed at \$24.95. 

This operational initiative contrasts with the advertising behavior
in RQ3. The unguided firm did not sustain advertising through weak
early returns; the guided firm did. In these runs, the agents developed
a concrete operating practice themselves, while the broader decision
to accept short-term losses for future growth still depended on
explicit strategic guidance.

\section{Conclusion}
\label{sec:conclusion}


MiniCorp provides a closed-loop office simulator that couples a persistent
agent firm with an evolving market. Its external world is grounded in
economic mechanisms and real marketplace data, with end-to-end fidelity
evaluated against empirical response patterns. As the firm and market
interact over time, MiniCorp generates enterprise records that connect
decisions to their business context and consequences. These records
also capture emergent organizational and operational behaviors developed
by the agents. Checkpointing allows the same situation to be replayed
under different decisions, adding counterfactual evidence unavailable
in static archives. Together, these features make MiniCorp a promising
source of scalable, realistic office-work data for evaluating and
training models on long-horizon organizational tasks. 
We envision MiniCorp as a scalable source of realistic office-work data for evaluating and training models on long-horizon organizational tasks. More broadly, the framework provides a principle that can be extended to additional firms, organizational structures, and industries.

\medskip

\bibliography{main}
\bibliographystyle{plainnat}

\appendix
\newpage
\section{Appendix}

\subsection{World Model Initializations and Agent Firm responses}
\begin{table*}[h]
\centering
\small
\begin{tabular}{p{0.20\textwidth}p{0.375\textwidth}p{0.235\textwidth}r}
\toprule
  \textbf{Situation} & \textbf{Initialized state or parameter} & \textbf{Route the world leaves open} & \textbf{Share} \\
\midrule
  A support backlog mixing genuine defects with false alarms & 16 aged claims on orders that arrived damaged and 16 on orders that arrived intact & Answer the genuine claims; refuse payment on the rest & 53.1\% \\
  Advertising that continues after the product has sold out & Inventory zeroed on the best-selling half of the catalogue, campaigns untouched & Cut the spend, restock, or both & 12.0\% \\
  A supplier shipping a high rate of defective units & Worst vendor's defect propensity raised to 25\% and bound to a third of the catalogue & Hold, audit, or replace the vendor & 7.6\% \\
  A seasonal dip in demand & Category demand pool cut by 40\%, leaving advertising efficiency untouched & None; the recovery is exogenous & 5.5\% \\
  Rivals bidding up the cost of advertising & Competing offers improved, so impressions cost more at unchanged click-through & Defend share; do not cut the budget & 3.3\% \\
  A product whose customer ratings have collapsed & One-star reviews seeded, lowering click-through at unchanged cost per impression & Repair the product's standing, not the budget & 2.8\% \\
  Four urgent problems and capacity for one & The highest-velocity product routed to the worst vendor, alongside three louder but self-limiting alarms & Work the compounding problem rather than the loudest & --- \\
  A product built to a cheap specification & One product's cell downgraded to the bottom tier, in the category weighting quality most heavily & Change the part & 0.4\% \\
  Listings switched off for a reason that no longer holds & Listings paused as out of stock, on well-rated products with stock on hand & Resume them & 0.2\% \\
  A shortage of one part used across many products & The component appearing in the most bills of material made scarce & Name the shared cause instead of treating each product separately & --- \\
\bottomrule
\end{tabular}
\caption{The ten situations, ordered by the share of archived actions submitted to the world model accounted for by the action types each one exercises. Each situation initializes selected state variables or latent parameters while preserving the structure of the world model.}
\label{tab:1}
\end{table*}

\begin{table*}[h]
\centering
\small
\begin{tabular}{p{0.14\textwidth}cp{0.25\textwidth}p{0.17\textwidth}p{0.21\textwidth}}
\toprule
\textbf{Situation} & \textbf{Found} & \textbf{What the company did} & \textbf{Collaboration} & \textbf{Why} \\
\midrule
Support backlog & 2/2 & Refunded the genuine claims, paid nothing on the false alarms & Irreplaceable, in part & Refund or replacement depends on a colleague's pending hold \\
Listings paused stale & 2/2 & Resumed all 6 listings in week 1 & Irreplaceable & The agent was not looking that way until a colleague wrote \\
Seasonal dip & 2/2 & Left advertising budgets and prices untouched & Unhelpful & Deadlock; right outcome anyway \\
Shared part shortage & 1/2 & Aimed purchase orders at the affected SKUs & Irreplaceable & The component call and the order sit in two roles \\
Advertising after sell-out & 1/2 & Paused all 15 sold-out listings once, none the second time & Helpful, then unhelpful & The second request was ignored \\
Cheap specification & 1/2 & Proposed and approved the component switch once & Helpful & --- \\
Rivals bidding up ads & 1/2 & Defended the budget once, cut it the second time & Absent & No message mentioned competitors \\
Ratings collapsed & 1/2 & Held the budget and worked the quality side & Unhelpful & Discussed widely, never handed off \\
Defective supplier & 0/2 & Audited in week 1 and held stock, but never replaced the vendor & Irreplaceable & The audit and the teardown sit in two roles \\
Four urgent problems & 0/2 & Worked the loudest alarm, not the compounding one & Unhelpful & A wrong priority was propagated \\
\bottomrule
\end{tabular}
\caption{Response to each situation, and what collaboration contributed.
\textbf{Found} counts the repetitions that reached the open route of Table~\ref{tab:1}.
\textbf{Collaboration} is audited chain by chain: \emph{irreplaceable} means the
agent could not have decided alone, \emph{helpful} means it happened and was
cited but the agent could have decided alone, \emph{unhelpful} means it happened
normally and did not produce the outcome, and \emph{absent} means it did not
happen.}
\label{tab:rq2}
\end{table*}

\end{document}